\documentclass[letterpaper]{article} % DO NOT CHANGE THIS
\usepackage{aaai2026}  % DO NOT CHANGE THIS
\usepackage{times}  % DO NOT CHANGE THIS
\usepackage{helvet}  % DO NOT CHANGE THIS
\usepackage{courier}  % DO NOT CHANGE THIS
\usepackage[hyphens]{url}  % DO NOT CHANGE THIS
\usepackage{graphicx} % DO NOT CHANGE THIS
\usepackage{natbib}  % DO NOT CHANGE THIS AND DO NOT ADD ANY OPTIONS TO IT
\usepackage{caption} % DO NOT CHANGE THIS AND DO NOT ADD ANY OPTIONS TO IT
\usepackage{algorithm}
\usepackage{algorithmic}
\usepackage{url}
\usepackage{amsmath}
\usepackage{amssymb}
\usepackage{mathtools}
\usepackage{ragged2e}
\usepackage{caption}
\usepackage{subcaption}
\usepackage{booktabs}
\usepackage{multirow}
\usepackage{makecell}
\usepackage{colortbl}
\usepackage{array}
\usepackage{diagbox}

\usepackage{booktabs}   % 更漂亮的表格线
\usepackage{multirow}
\usepackage{tabularx}   % 自动列宽
\usepackage{siunitx}    % 数字对齐
\newcommand\ChangeRT[1]{\noalign{\hrule height #1}}
\usepackage{newfloat}
\usepackage{listings}
\DeclareCaptionStyle{ruled}{labelfont=normalfont,labelsep=colon,strut=off} % DO NOT CHANGE THIS
\floatstyle{ruled}
\newfloat{listing}{tb}{lst}{}
\floatname{listing}{Listing}
\title{Sparse4DGS: 4D Gaussian Splatting for Sparse-Frame Dynamic \\ Scene Reconstruction}

\author{
    Changyue Shi \textsuperscript{\rm 1,2}\thanks{Project Leader: changyue\_shi@163.com.}, 
    Chuxiao Yang \textsuperscript{\rm 1}, 
    Xinyuan Hu \textsuperscript{\rm 1}, 
    Minghao Chen \textsuperscript{\rm 1}, 
    Wenwen Pan \textsuperscript{\rm 1}, \\
    Yan Yang \textsuperscript{\rm 1}, 
    Jiajun Ding \textsuperscript{\rm 1} \thanks{Corresponding Author: djj@hdu.edu.cn.},
    Zhou Yu \textsuperscript{\rm 1}, 
    Jun Yu \textsuperscript{\rm 3}
}
\affiliations {
    \textsuperscript{\rm 1}Hangzhou Dianzi University,
    \textsuperscript{\rm 2}Peking University,
    \textsuperscript{\rm 3}Harbin Institute of Technology 
}
\usepackage{bibentry}
\begin{document}
\maketitle

\begin{figure*}[t]
\centering
\includegraphics[width=1.0\textwidth]{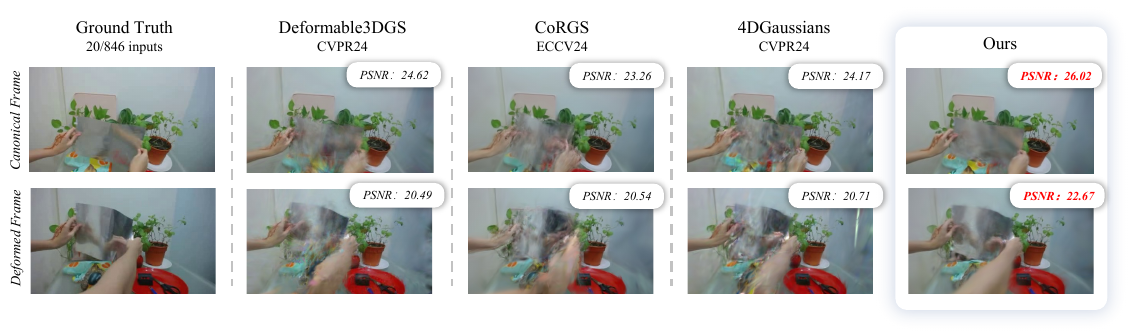} % Reduce the figure size so that it is slightly narrower than the column.
\caption{In this work, we introduce Sparse4DGS, a novel approach for dynamic scene reconstruction using sparse input frames. In the ``\textit{Sheet}'' scene from the NeRF-DS~\cite{yan2023nerf} dataset, when taking sparse frames as inputs, Sparse4DGS achieves high-quality novel view synthesis results in both canonical and deformed spaces.}
\label{fig:teaser}
\end{figure*}

\begin{abstract}
Dynamic Gaussian Splatting approaches have achieved remarkable performance for 4D scene reconstruction. However, these approaches rely on dense-frame video sequences for photorealistic reconstruction. In real-world scenarios, due to equipment constraints, sometimes only sparse frames are accessible. In this paper, we propose \textbf{Sparse4DGS}, the first method for sparse-frame dynamic scene reconstruction. 
We observe that dynamic reconstruction methods fail in both canonical and deformed spaces under sparse-frame settings, especially in areas with high texture richness. Sparse4DGS tackles this challenge by focusing on texture-rich areas. 
For the deformation network, we propose Texture-Aware Deformation Regularization,  which introduces a texture-based depth alignment loss to regulate Gaussian deformation. For the canonical Gaussian field, we introduce Texture-Aware Canonical Optimization, which incorporates texture-based noise into the gradient descent process of canonical Gaussians. 
Extensive experiments show that when taking sparse frames as inputs, our method outperforms existing dynamic or few-shot techniques on NeRF-Synthetic, HyperNeRF, NeRF-DS, and our iPhone-4D datasets.
\end{abstract}

\begin{links}
    \centering
    \link{Project Page}{ChangyueShi.github.io/Sparse4DGS}
    % \link{Datasets}{https://aaai.org/example/datasets}
    % \link{Extended version}{https://aaai.org/example/extended-version}
\end{links}

\section{Introduction}
\label{sec:intro}

Modeling 4D scenes from 2D images and synthesizing photorealistic novel views is essential for computer vision and graphics~\cite{pumarola2021d, yang2024deformable, yang2023real}. This task has attracted significant attention from both industry and academia due to its great potential in augmented and virtual reality applications~\cite{yang20244d, jiang2024vr}. The recent emergence of 3D Gaussian Splatting~\cite{kerbl3Dgaussians} has revolutionized dynamic scene reconstruction through its real-time rendering capabilities. 

To reconstruct a dynamic scene, a variety of Gaussian Splatting methods~\cite{huang2024sc, yang2024deformable, liang2023gaufre, duisterhof2023md, wu20244d, duan20244d, yang2023real} have been proposed. The most commonly used approach explicitly represents the scene as a canonical Gaussian field and models Gaussian variations at specific timestamps using a deformation network~\cite{yang2024deformable, wu20244d}. However, existing methods depend on dense-frame video sequences as input to reconstruct high-fidelity dynamic scenes. In real-world applications, sometimes, only low-FPS video sequences are available due to device limitations.

This naturally raises the question: \textbf{Can high-quality 4D scenes be reconstructed from sparse frames?}
We investigate the capability of 4D Gaussian Splatting for addressing sparse-frame inputs and demonstrates that it is possible to reconstruct a high-fidelity dynamic scene even from a limited sequence of frames. When the input image frames are sparse, existing methods~\cite{yang2024deformable, wu20244d} suffer a significant degradation. 
We observe that this degradation manifests in texture‑rich regions, as these areas contain abundant high‑frequency content that is challenging to preserve under deformation. In these areas, the geometry tends to collapse in deformed and canonical spaces, as shown in the second and fourth columns of Fig.~\ref{fig:teaser}. 

In this work, we propose \textbf{Sparse4DGS}, a novel framework tailored for dynamic scene reconstruction from sparse-frame video sequences. Our intuition is that sparse-frame inputs inherently provide limited information, in which scenario, high-frequency texture signals become the primary source of rich detail and dynamic cues essential for accurate deformation modeling. Therefore, we encourage Gaussians to focus on these texture-rich regions, thereby modeling the underlying structure.  Central to our approach is the \textit{Texture Intensity (TI) Gaussian Field}: for each input frame, we compute per-pixel gradient magnitudes to derive 2D \textit{TI} maps. We then embed this texture intensity into each 3D Gaussian by aligning rendered and extracted 2D \textit{TI} maps.

We tackle the problems from the perspective of texture intensity consistency in both deformed and canonical spaces.
For the Gaussian deformation, we propose \textit{Texture-Aware Deformation Regularization (TADR)}. Apart from aligning the \textit{TI} maps, we incorporate a texture-aware depth regularization to guide the geometric structure of the deformed space. Specifically, we extract texture intensity from depth maps to obtain \textit{TI} maps of depth. The texture of rendered depth at a specific timestamp is aligned with the depth texture map extracted from the depth map predicted by Mono-Depth Estimator~\cite{ranftl2021vision}. For the canonical Gaussians, we propose \textit{Texture-Aware Canonical Optimization (TACO)}. In \textit{TACO}, the gradient descent of canonical Gaussians is reformulated based on Stochastic Gradient Langevin Dynamics~\cite{welling2011bayesian, brosse2018promises, kheradmand2024accelerating, kheradmand20243d} with a noise term based on the value of \textit{TI} attribute. This noise term continuously perturbs the optimization of canonical Gaussians until they converge to texture-rich areas. When using inputs with either 5 FPS (low frame rate) or 30 FPS (high frame rate) for training, Sparse4DGS exhibits superior dynamic reconstruction, demonstrating its potential for a wide range of frame rate videos in real-world applications.  

In summary, the major contributions of our work are:
\begin{itemize}
    \item We demonstrate that high-quality dynamic scenes can be reconstructed from sparse-frame video sequences. To the best of our knowledge, this is the first work to focus on sparse-frame 4D scene reconstruction. 
    
    \item We tackle the problems of optimization under sparse inputs from the perspective of high-frequent texture information and propose a texture-aware method (\textit{TADR} and \textit{TACO}) to align Gaussians with texture richness, thus preserving the underlying geometric structure.

    \item Extensive experiments demonstrate that Sparse4DGS outperforms previous dynamic and sparse NVS methods across multiple datasets, including NeRF-Synthetic, HyperNeRF, NeRF-DS, and our iPhone-4D dataset.
 
\end{itemize}

\begin{figure*}[t]
    \centering
    \includegraphics[width=0.9\linewidth]{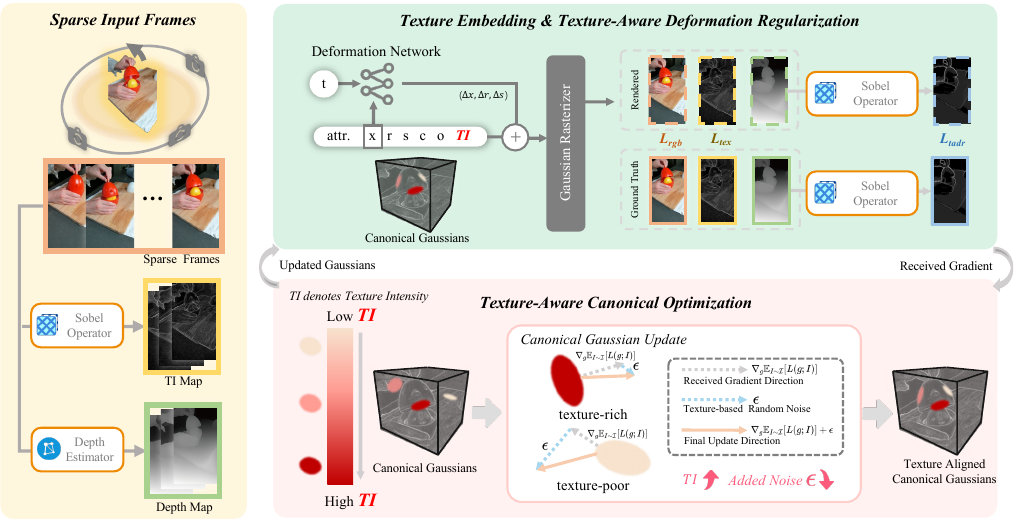}
    \caption{\textbf{Overall pipeline of Sparse4DGS.} 
    \textit{Left:} The Sobel operator and Mono-Depth Estimator \cite{ranftl2021vision} are employed to generate the texture intensity (TI) and depth maps from sparse frames.
    \textit{Top right:} The $TI$ attribute is embedded in each Gaussian via $L_{tex}$. \textit{Texture-Aware Deformation Regularization} is employed to align the rendered and ground truth texture intensity of depth maps with $L_{tadr}$. 
    \textit{Bottom right:} After receiving the original gradient, \textit{Texture-Aware Canonical Optimization} introduces an additional texture-based noise to each Gaussian, thereby improving their concentration on texture-rich regions.
    % Based on Deformable3DGS~\cite{yang2024deformable}, we propose a novel texture-based depth regularization method by minimizing the relevancy $L_{tadr}$ between the rendered and ground truth textures of depth maps. Meanwhile, the \textit{Texture} value of each Gaussian is optimized through the minimization of $L_{tex}$ that aligns the ground truth and rendered texture maps. \textit{Bottom left:} After receiving the original gradient, additional noise is added to each Gaussian. The noise depends on the \textit{Texture (Tex)} value of each Gaussian, with texture-poor Gaussians receiving greater noise than texture-rich ones. \textit{Right:} The Texture Extractor (TE) and Dense Prediction Transformer (DPT) \cite{ranftl2021vision} are employed to generate the texture and depth maps from limited input RGB frames.
    } 
    \label{fig:pipeline}
\end{figure*}

\section{Related Work}
\noindent \textit{\textbf{Novel View Synthesis}}
Novel view synthesis (NVS) aims to generate unseen views of a scene from a set of input images. Traditional methods, such as Structure-from-Motion (SfM)~\cite{ullman1979interpretation} and Multi-View Stereo (MVS)~\cite{tomasi1992shape}, rely on geometric reconstruction techniques. Neural Radiance Field (NeRF)~\cite{mildenhall2021nerf} introduces a learning-based approach, representing a scene as a volumetric radiance field parameterized by an MLP. While subsequent NeRF-based methods~\cite{muller2022instant, chen2022tensorf, yu2021plenoxels, barron2021mip, barron2022mip} have enhanced rendering quality and efficiency of the vanilla NeRF, they remain constrained by the computational overhead of volumetric rendering.
Recently, Gaussian Splatting~\cite{kerbl3Dgaussians} has emerged as an efficient alternative, leveraging rasterization to achieve real-time, high-fidelity scene reconstruction. It has inspired extensive researches across various domains, including dynamic scene reconstruction~\cite{wu20244d, yang2024deformable, huang2024sc}, few-shot learning~\cite{zhu2025fsgs, li2024dngaussian, zhang2025cor, paliwal2025coherentgs}, and super-resolution reconstruction~\cite{feng2024srgs}, among others~\cite{shi2025realm}.

\noindent \textit{\textbf{Gaussian Splatting for Dynamic Scenes}}
Many studies have extended Gaussian Splatting to dynamic scene reconstruction.
 Research in this area can be broadly divided into two categories: deformation-based and 4D Gaussian-based methods. Deformation-based methods use an implicit structure~\cite{huang2024sc, yang2024deformable, liang2023gaufre, duisterhof2023md, wu20244d} to model the deformation from a static canonical Gaussian field. On the other hand, 4D Gaussian-based methods~\cite{duan20244d, yang2023real} introduce the time dimension in the original Gaussian Splatting formulation, which can be expressed with the 4D position and 4D covariance matrix. Many studies have now applied dynamic Gaussian Splatting to various fields. ZDySS~\cite{saroha2025zdyss} proposes a zero-shot scene stylization method with dynamic Gaussian Splatting. Some methods~\cite{shan2025deformable} utilize it for reconstruction of surgical scenes. Deblur4DGS~\cite{wu2024deblur4dgs} proposes a framework in dynamic Gaussian Splatting for blurry monocular video. Existing dynamic Gaussian Splatting methods rely on dense-frame video sequences as input for dynamic scene reconstruction. In this work, we propose Sparse4DGS and show that high-quality dynamic scenes can be reconstructed using sparse input frames.

\noindent \textit{\textbf{Few Shot Gaussian Splatting}}
Gaussian Splatting~\cite{kerbl3Dgaussians} has gained significant attention for its fast rendering speed and high-quality results across various applications~\cite{feng2024srgs, qin2024langsplat, lee2025deblurring, gao2025relightable, choi2025click, shi2025mmgs, hu2025texture}. Despite these advantages, novel view synthesis (NVS) still requires hundreds of input images to achieve photo-realistic reconstructions, limiting its practicality in real-world scenarios. To address this, recent research has explored techniques for high-quality reconstruction from sparse inputs. GaussianObject~\cite{yang2024gaussianobject} leverages a diffusion model~\cite{rombach2021highresolution} to synthesize photo-realistic novel views from just four input images. DNGaussian~\cite{li2024dngaussian} introduces a depth regularization module that enhances reconstruction efficiency and quality. CoRGS~\cite{zhang2025cor} mitigates reconstruction inaccuracies by refining the training process. FSGS~\cite{zhu2025fsgs} tackles the challenge of extremely sparse SfM initialization by introducing a Gaussian Unpooling mechanism. CoherentGS~\cite{paliwal2025coherentgs} proposes a structured Gaussian representation that enables explicit control in 2D image space. Additionally, generative approaches~\cite{chen2025mvsplat, charatan2024pixelsplat} have been explored to predict Gaussians directly using feed-forward networks, further improving performance.

\begin{figure}
    \centering
    \includegraphics[width=0.80\linewidth]{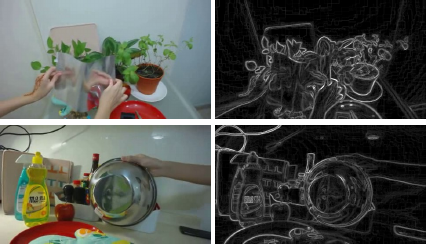}
   \caption{\textbf{Visualization of texture intensity maps.} \textit{Left:} Input RGB images. \textit{Right:} Extracted texture intensity maps.}
    \label{fig:vis_texture}
\end{figure}

\section{Preliminary}
\subsection{Dynamic Gaussian Splatting}  
Gaussian Splatting~\cite{kerbl3Dgaussians} represents scenes with anisotropic Gaussians $\{ \mu_i, s_i, r_i, o_i, c_i \}$, where $\mu_i \in \mathbb{R}^3$ is the position, $s_i \in \mathbb{R}^3$ and $r_i \in \mathbb{R}^4$ define the covariance, $o_i \in \mathbb{R}$ is opacity, and $c_i \in \mathbb{R}^K$ encodes color. The rendering process follows alpha-blending:  
\begin{equation}  
\label{eq:render}
c=\sum_{i=1}^{n}c_{i}\alpha _{i}\prod_{j=1}^{i-1}(1-\alpha _{j}),
\end{equation}  
where $\alpha_i$ is computed by projecting Gaussians to the image plane.
To model dynamic scenes, recent methods \cite{yang2024deformable, wu20244d} extend it with a deformation field. Given time $t$, the deformation MLP predicts offsets for each Gaussian:
\begin{equation}  
(\delta {x}, \delta {r}, \delta {s}) = \mathcal{F_\theta}(\gamma(\operatorname{sg}({x})), \gamma(t)),
\end{equation}  
where $\mathcal{F_\theta}$ is MLP, $\gamma(\cdot)$ denotes positional encoding \cite{mildenhall2021nerf}, and $\operatorname{sg}(\cdot)$ stops gradient propagation. The deformed Gaussians $\{ \mu_i{+}\delta x, s_i{+}\delta s, r_i{+}\delta r, o_i, c_i \}$ are rendered via Eq.~\ref{eq:render}. An RGB loss is utilized to optimize the canonical Gaussians and the deformation network:
\begin{equation}
    \label{3dgs_loss}
    L_{rgb}=(1-\lambda )L_{1}(\hat{I},I)+\lambda L_{SSIM}(\hat{I},I),
\end{equation}
where $\hat{I}$ is the rendered image and $I$ is the corresponding ground truth (GT). Here, $L_1$ represents the Mean Squared Error (MSE) loss, while $L_{SSIM}$ is the Structural Similarity Index Measure (SSIM) \cite{wang2004image}.

\subsection{Stochastic Gradient Langevin Dynamics}
The Stochastic Gradient Langevin Dynamics (SGLD) \cite{welling2011bayesian, brosse2018promises} method has been recently applied to novel view synthesis applications \cite{kheradmand2024accelerating, kheradmand20243d}. The traditional SGD update rule \cite{amari1993backpropagation} in the original Gaussian Splatting can be described as:
\begin{equation}
\label{eq:sgd}
    {g} = {g} - \alpha \cdot \nabla_{{g}} \mathbb{E}_{{I} \sim \mathcal{I}} \left[ L({g}; {I}) \right],
\end{equation}
where $\alpha$ is the learning rate, $\mathbb{E}_{{I} \sim \mathcal{I}}$ represents the expectation over the data distribution $\mathcal{I}$, and $L({g}; {I})$ is the total loss function that depends on the parameter ${g}$ and the input image ${I}$. Meanwhile, the SGLD update takes the form as:
\begin{equation}
    {g} = {g} + a \cdot \nabla_{{g}} \log \mathcal{P}({g}) + b \cdot {\epsilon},
\end{equation}
where $\mathcal{P}$ is the data-dependent probability density function of the target distribution. $\boldsymbol{\epsilon}$ represents the noise distribution for exploration. Hyperparameters $a$ and $b$ control the trade-off between convergence speed and exploration. In Gaussian Splatting application~\cite{kheradmand20243d}, it can be rewritten as:
\begin{equation}
    {g} = {g} - \alpha_{g} \cdot \nabla_{{g}} \mathbb{E}_{{I} \sim \mathcal{I}} \left[ L({g}; {I}) \right] + \alpha_{noise} \cdot \epsilon_{o},
\end{equation}
where the noise term $\epsilon_o$ can be described as follows:
\begin{equation}
    \label{eq:noise_opa}
    \epsilon_{o} = \sigma(-k(o-t)) \cdot \sum \eta.
\end{equation}

Here, $\eta \sim \mathcal{N}(0, I)$, and $k$ and $t$ are set to 100 and 0.995, respectively, to create a sharp transition function that shifts from zero to one. $\sigma$ represents the sigmoid function and $o$ is the opacity value of each Gaussian. In this work, we further extend this formulation to better suit dynamic scene reconstruction under sparse inputs. 

\begin{figure}
    \centering
    \includegraphics[width=0.90\linewidth]{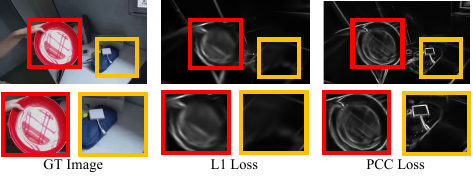}
   \caption{\textbf{Comparison between L1 distance and PCC for the rendered texture map.} ${L}_{tex}$ with PCC achieves more precise texture embedding results.}
    \label{fig:edge_pcc_l1}
\end{figure}

\begin{table*}[t] 
\centering
\scriptsize

\begin{tabular}{l|ccc|ccc|ccc|ccc|ccc}
\multirow{2}{*}{Methods} & \multicolumn{3}{c|}{NeRF-Synthetic (20 frames)} & \multicolumn{3}{c|}{NeRF-DS (20 frames)} & \multicolumn{3}{c|}{Hyper-NeRF (30 frames)} & \multicolumn{3}{c|}{iPhone-4D (30 FPS)} & \multicolumn{3}{c}{iPhone-4D (5 FPS)} \\ 
\cline{2-16} & PSNR & SSIM & LPIPS & PSNR & SSIM & LPIPS & PSNR & SSIM & LPIPS & PSNR & SSIM & LPIPS & PSNR & SSIM & LPIPS \\ 
\ChangeRT{1.2pt}
Deformable3DGS & 22.65 & 0.927 & 0.073 & 20.81 & 0.753 & 0.301 & 22.41 & 0.661 & 0.295 & 27.01 & 0.909 & 0.210 & 21.12 & 0.817 & 0.299 \\
4DGaussians & 22.47 & 0.931 & 0.071 & 19.70 & 0.697 & 0.350 & 20.64 & 0.637 & 0.414 & 28.79 & 0.896 & 0.246 & 16.37 & 0.678 & 0.421 \\
CoRGS & 20.15 & 0.920 & 0.089 & 19.86 & 0.746 & 0.319 & 20.50 & 0.638 & 0.364 & 21.58 & 0.851 & 0.266 & 16.81 & 0.737 & 0.374 \\
\textbf{Ours} & \textbf{25.31} & \textbf{0.944} & \textbf{0.056} & \textbf{22.34} & \textbf{0.801} & \textbf{0.233} & \textbf{23.91} & \textbf{0.711} & \textbf{0.294} & \textbf{29.81} & \textbf{0.929} & \textbf{0.177} & \textbf{27.51} & \textbf{0.910} & \textbf{0.205} \\ 
\end{tabular} 
\caption{\textbf{Quantitative comparison on multiple datasets.} The metrics include PSNR $\uparrow$, SSIM $\uparrow$, and LPIPS $\downarrow$. Best results are highlighted in \textbf{bold}.} 
\label{tab:comparison_all} 
\end{table*}

% \begin{figure*}[t]
%     \centering
%     \includegraphics[width=0.94\linewidth]{Figs/vis_hyper.pdf}
%     \caption{\textbf{Visualization on Hyper-NeRF dataset with 30 inputs.} Sparse4DGS
% outperforms baseline methods.} 
%     \label{fig:vis_hyper_30}
% \end{figure*}

\begin{figure*}
    \centering
    \includegraphics[width=0.9\linewidth]{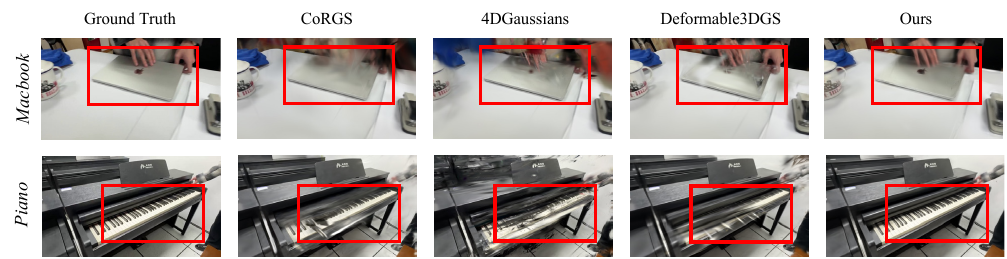}
    \caption{\textbf{Visualization of various methods on our iPhone-4D dataset with 5 FPS video inputs.}} 
    \label{fig:vis_iphone_50}
\end{figure*}
\section{Proposed Method}
As illustrated in Fig.~\ref{fig:pipeline}, we propose Sparse4DGS. In Sec.~\ref{sec:texture_extractor}, we introduce \textit{Texture Intensity Gaussian Field}.
In Sec.~\ref{sec:tadr}, we introduce
\textit{Texture-Aware Deformation Regularization (TADR)}. 
In Sec.~\ref{sec:taco}, we introduce the \textit{Texture-Aware Canonical Optimization (TACO)}.

\subsection{Representing 3D Texture Richness}
\label{sec:texture_extractor}
To encourage texture consistency, we first propose the Texture Intensity Gaussian Field. We extract texture maps from input frames using a discrete differential operator. The texture information is then embedded into each Gaussian.\noindent \textbf{\textit{2D Texture Intensity Maps.}} 
We represent texture richness using per‑pixel gradients. These gradient values quantify the variation in colors between adjacent pixels and explicitly capture local texture richness. 

Given an input RGB image \(I\in\mathbb{R}^{H\times W\times3}\), we first compute two gradient maps by convolving \(I\) with the standard horizontal and vertical Sobel operators, respectively:
\begin{equation}
\begin{aligned}
TI_x = I \ast 
\begin{bmatrix}
-1 &  0 & 1\\
-2 &  0 & 2\\
-1 &  0 & 1
\end{bmatrix},
\, TI_y = I \ast 
\begin{bmatrix}
-1 & -2 & -1\\
 0 &  0 &  0\\
1 & 2 & 1
\end{bmatrix}.
\end{aligned}
\end{equation}

We then compute the per‑pixel gradient magnitude:

\begin{equation}
\begin{aligned}
TI_{gt}(i,j) &= \sqrt{TI_x(i,j)^2 + TI_y(i,j)^2},
\end{aligned}
\end{equation}
where \(TI(i,j)\) is the gradient magnitude at pixel \((i,j)\), which serves as an explicit measure of local texture richness. Fig.~\ref{fig:vis_texture} presents the extracted texture intensity maps.

\noindent\textit{\textbf{Texture Intensity in 3D Space.}} To represent the texture richness in 3D space, we introduce a new attribute \textit{Texture Intensity $(TI)$} for each Gaussian. $TI$ can be rendered into a texture map $TI_{render}$ via differentiable rasterizer as shown in Eq.~\ref{eq:render}. 
Since the texture extractor is applied independently to each image, spatial inconsistencies may occur. The traditional L1 loss primarily focuses on absolute differences between pixels, disregarding spatial inconsistencies. To relax this issue, Sparse4DGS employs the Pearson Correlation Coefficient (PCC)~\cite{cohen2009pearson} to compute the relative variation rates between $TI_{gt}$ and $TI_{render}$. The formulation of PCC can be decribed as:
\begin{equation}
\label{eq:pcc_equation}
\text{PCC}(X, Y) = \frac{Cov(X, Y)}{\sqrt{Var(X)} \cdot \sqrt{Var(Y)}},
\end{equation}
where $X$ and $Y$ represent two distinct variables, $Cov(X, Y)$ denotes their covariance, and $Var(X)$ and $Var(Y)$ represent their variances. We optimize $TI$ through the PCC loss between $TI_{gt}$ and $TI_{render}$:
\begin{equation}
\label{eq:l_tex}
{L}_{tex} = 1 - \text{PCC}(TI_{gt} , TI_{render}).
\end{equation} 

This approach relaxes spatial inconsistencies and achieves superior texture embedding results (Fig.~\ref{fig:edge_pcc_l1}).

\subsection{Texture-Aware Deformation Regularization}
\label{sec:tadr}

The optimization of Sparse4DGS builds upon previous dynamic Gaussian Splatting method~\cite{yang2024deformable}. The optimization performance under sparse frames is primarily influenced by the deformation network and the canonical Gaussian field. This section focuses on the regularization of the deformation field. We propose \textit{Texture-Aware Deformation Regularization (TADR)} to guide the geometric structure of deformed Gaussians with depth-texture consistency.

Texture intensity is often correlated with changes in depth. For efficient geometric constraint, a previous method \cite{zhu2025fsgs} computes the PCC between the rendered depth maps and the monocular depth estimated by DPT~\cite{ranftl2021vision}. This image-level regularization has limitations in capturing local depth variations (Fig.~\ref{fig:tadr_l1}). To efficiently constrain the deformation network, Sparse4DGS develops a texture-based regularization method. 

To begin with, TADR utilizes Sobel (Sec.~\ref{sec:texture_extractor}) to extract texture intensity maps information of the rendered depth $D_{render}$ and depth computed from DPT~\cite{ranftl2021vision} $D_{dpt}$: 
$TI_{render}^{depth} = \text{TE}(D_{render}),TI_{gt}^{depth}= \text{TE}(D_{dpt}).$ Therefore, the \textit{TADR} loss can be formulated as:
\begin{equation}
\label{eq:l_tadr}
{L}_{tadr} = 1 - \text{PCC}(TI_{gt}^{depth} , TI_{render}^{depth}).
\end{equation} 

Similar to $L_{tex}$, \textit{TADR} employs PCC for supervision, since $TI_{gt}^{depth}$ exhibits spatial inconsistency. Therefore, the overall training loss can be described as: 
\begin{equation}
\label{eq:overall_loss}
    L=L_{rgb} + \lambda_{1} \cdot L_{tex} + \lambda_{2} \cdot L_{tadr},
\end{equation}
where $L_{rgb}$ consists of an MSE loss and an SSIM loss~\cite{kerbl3Dgaussians}, as shown in Eq.~\ref{3dgs_loss}.

\subsection{Texture-Aware Canonical Optimization}
\label{sec:taco}
The supervision defined in Eq.~\ref{eq:overall_loss} generates a per-Gaussian gradient $\nabla_{{g}} \mathbb{E}_{{I} \sim \mathcal{I}} \left[ L({g}; {I}) \right]$, which drives the canonical Gaussian to update its attributes using the SGD method, as formalized in Eq.~\ref{eq:sgd}. To help canonical Gaussian field concentrate more on texture-rich areas, we reformulate this update function based on SGLD~\cite{welling2011bayesian,kheradmand2024accelerating, kheradmand20243d} and propose \textit{Texture-Aware Canonical Optimization (TACO)}.

The main idea of \textit{TACO} is to introduce stochastic noise in each iteration during the Gaussian update. Our update formulation for the canonical Gaussians can be described as:
\begin{equation}
{g} = {g} - \alpha_{g} \cdot \nabla_{{g}} \mathbb{E}_{{I} \sim \mathcal{I}} \left[ L({g}; {I}) \right] + \alpha_{noise} \cdot (\epsilon_{tex}+\epsilon_{o}),
\end{equation}
where the $\epsilon_o$ is employed to reduce ambiguous Gaussians, as described in Eq.~\ref{eq:noise_opa}, which often manifest as floaters. \textit{TACO} introduces an additional texture aware term
$\epsilon_{tex}$, which is defined as:
$\epsilon_{tex} = \sigma(-k(TI-t)) \cdot \sum \eta.$
Since the value range of $TI$ is identical to that of $opacity$, we use the same hyperparameters as those in Eq.~\ref{eq:noise_opa}. A Gaussian with a greater $TI$ value would receive a smaller noise term. When a Gaussian reaches texture-rich regions, its $TI$ value approaches one, and $\epsilon_{tex}$ approaches zero. Consequently, $\epsilon_{tex}$ continuously influences the optimization process until the Gaussians attain high $opacity$ and $Tex$ values, indicating that they have reached texture-rich regions.

\begin{figure*}[t]
    \centering   
    \includegraphics[width=0.85\linewidth]{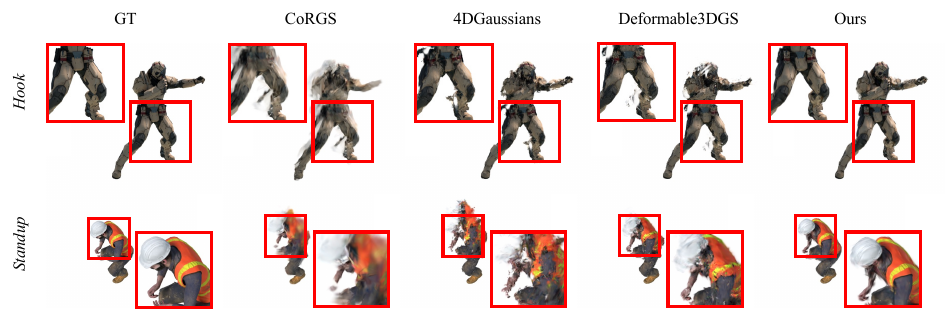}
    \caption{\textbf{Qualitative results on the NeRF-Synthetic dataset with 20 inputs.} The results of various dynamic reconstruction methods are demonstrated from new viewpoints and also in new frames.} 
    \label{fig:vis_blender_20}
\end{figure*}

\begin{figure*}
    \centering
    \includegraphics[width=0.85\linewidth]{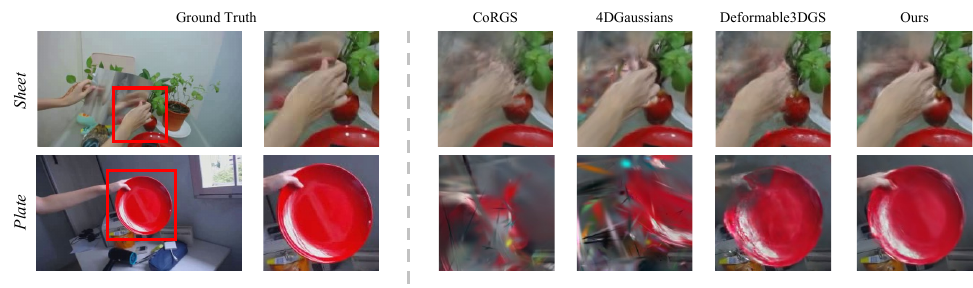}
    \caption{\textbf{Qualitative results on the NeRF-DS dataset with 20 inputs.} Our method consistently
outperforms other baselines.} 
    \label{fig:vis_nerfds_20}
\end{figure*}

\begin{table*}[t]
\scriptsize
\centering

% 第一行 (3 个子表)
\begin{subtable}[t]{0.32\textwidth}
    \centering
    \begin{tabular}{cccc}
        Methods & PSNR & SSIM & LPIPS  \\ 
        \ChangeRT{1.2pt}
        Baseline & 20.81 & 0.753 & 0.301 \\ 
        w/o TADR & 21.89 & 0.792 & 0.245 \\ 
        w/o TACO & 21.33 & 0.773 & 0.271 \\ 
        Ours & \textbf{22.34} & \textbf{0.801} & \textbf{0.233} \\ 
    \end{tabular}
    \captionsetup{font=footnotesize}
    \subcaption{\textbf{Ablation Study on TADR and TACO.}}
    \label{tab:major_ablation}
\end{subtable}
\quad
\begin{subtable}[t]{0.32\textwidth}
    \centering
    \begin{tabular}{cccc}
        $L_{tex}$ weight & PSNR & SSIM & LPIPS \\ 
        \ChangeRT{1.2pt}
        $\lambda_1=0.001$ & 21.80 & 0.789 & 0.249 \\ 
        $\lambda_1=0.01$ & \textbf{22.34} & \textbf{0.801} & \textbf{0.233} \\ 
        $\lambda_1=0.1$ & 21.79 & 0.788 & 0.254 \\ 
        $\lambda_1=1$ & 20.17 & 0.714 & 0.353 \\ 
    \end{tabular}
    \captionsetup{font=footnotesize}
    \subcaption{\textbf{Ablation study on $L_{tex}$ weight $\lambda_1$.}}
    \label{tab:lambda_1}
\end{subtable}
\quad
\begin{subtable}[t]{0.32\textwidth}
    \centering
    \begin{tabular}{cccc}
        $L_{tadr}$ weight & PSNR & SSIM & LPIPS  \\ 
        \ChangeRT{1.2pt}
        $\lambda_2=0.001$ & 22.04 & 0.797 & 0.240 \\ 
        $\lambda_2=0.01$ & \textbf{22.34} & \textbf{0.801} & \textbf{0.233} \\ 
        $\lambda_2=0.1$ & 22.12 & 0.797 & 0.241 \\ 
        $\lambda_2=1$ & 21.86 & 0.794 & 0.243 \\ 
    \end{tabular}
    \captionsetup{font=footnotesize}
    \subcaption{\textbf{Selection of hyper-parameter $\lambda_2$.}}
    \label{tab:lambda_2}
\end{subtable} \\[1em]

% 第二行 (3 个子表)
\begin{subtable}[t]{0.32\textwidth}
    \centering
    \begin{tabular}{cccc}
        Methods & PSNR & SSIM & LPIPS  \\ 
        \ChangeRT{1.2pt}
        TACO w/o $\epsilon_o$ & 21.81 & 0.792 & 0.246 \\ 
        TACO w/o $\epsilon_{tex}$ & 21.57 & 0.783 & 0.260 \\ 
        Ours & \textbf{22.34} & \textbf{0.801} & \textbf{0.233} \\ 
    \end{tabular}
    \captionsetup{font=footnotesize}
    \subcaption{\textbf{Ablation study on $\epsilon_o$ or $\epsilon_{tex}$ in TACO.}}
    \label{tab:taco_epsilon}
\end{subtable}
\quad
\begin{subtable}[t]{0.32\textwidth}
    \centering
    \begin{tabular}{cccc}
        Methods & PSNR & SSIM & LPIPS  \\ 
        \ChangeRT{1.2pt}
        $L_{tex}$ w/o PCC & 21.71 & 0.789 & 0.245 \\ 
        $L_{tadr}$ w/o PCC & 22.09 & 0.797 & 0.239 \\ 
        Ours & \textbf{22.34} & \textbf{0.801} & \textbf{0.233} \\ 
    \end{tabular}
    \captionsetup{font=footnotesize}
    \subcaption{\textbf{PCC loss (replaced with L1 loss).}}
    \label{tab:pcc_ablate}
\end{subtable}
\quad
\begin{subtable}[t]{0.32\textwidth}
    \centering
    \begin{tabular}{cccc}
        Methods & PSNR & SSIM & LPIPS  \\ 
        \ChangeRT{1.2pt}
        w/o texture-aware & 21.46 & 0.775 & 0.277 \\ 
        Ours & \textbf{22.34} & \textbf{0.801} & \textbf{0.233} \\ 
        & & & \\ 
    \end{tabular}
    \captionsetup{font=footnotesize}
    \subcaption{\textbf{Texture-aware depth loss in TADR.}}
    \label{tab:tadr_detail}
\end{subtable}

\caption{\textbf{Ablation study.} We ablate our method on NeRF-DS dataset with 20 input views. The evaluation metrics include PSNR $\uparrow$, SSIM $\uparrow$, and LPIPS $\downarrow$. Best metrics are highlighted in \textbf{bold}.}
\label{tab:abla}
\end{table*}

\section{Experiments}
\label{sec:experiment}
\subsection{Experimental Settings}
\textbf{Dynamic datasets.}
The NeRF-Synthetic dataset \cite{pumarola2021d} contains 8 synthetic scenes. 
%We evaluate Sparse4DGS with 20, 30, and 40 input frames, which are uniformly sampled from the original training set. 
The NeRF-DS dataset~\cite{yan2023nerf} consists of 7 real-world scenes, each containing images captured using two monocular cameras. For NeRF-Synthetic and NeRF-DS datasets, we uniformly samples 20, 30, and 40 frames from their original training sets for model training and use the same evaluation sets (keeping the original number of frames) as previous works~\cite{yang2024deformable}. The Hyper-NeRF dataset~\cite{park2021hypernerf} includes 17 real-world scenes, with 5 typical scenes selected in this work for comparison. For Hyper-NeRF dataset, we uniformly samples 10, 20, and 30 frames from the original training set and adopt the same evaluation set as previous methods~\cite{yang2024deformable}.

\noindent\textbf{iPhone-4D dataset.} To demonstrate the capability of Sparse4DGS in various real-world frame-rate application, we introduce the iPhone-4D dataset, which consists of four real-world scenes. This dataset is captured using an iPhone at 30 FPS. We evaluate our method on this dataset using training inputs at 5 FPS and 30 FPS. The selection of evaluation set follows Hyper-NeRF dataset. We provide details of this dataset in the supplementary materials.

\noindent\textbf{Baselines and metrics.} We compare Sparse4DGS with typical dynamic and few-shot scene reconstruction methods including Deformable3DGS (CVPR24)~\cite{yang2024deformable}, 4DGaussians (CVPR24)~\cite{wu20244d}, and CoRGS (ECCV24)~\cite{zhang2025cor}. We run these methods on four datasets without modifying their original settings. In the following sections, we report the average PSNR, SSIM~\cite{wang2004image}, and LPIPS~\cite{zhang2018unreasonable}.
% \noindent\textbf{Implementation.} We implemented Sparse4DGS using PyTorch framework. $\lambda_{1}$ and $\lambda_{2}$ are set to 0.01 as default. The total optimization time is approximately 20 minutes for real-world datasets and 15 minutes for synthetic datasets. The camera poses and initial point clouds are generated using Colmap. Results can be obtained using a NVIDIA RTX 3090 GPU.

\subsection{Main Results with Sparse Inputs}
In this section, we evaluate our method and baselines using sparse-frame inputs to explore the potential of Sparse4DGS. 

\noindent\textbf{Quantitative results.} 
 A subset of the quantitative results is presented in Tab.~\ref{tab:comparison_all}. For most datasets, our method improves PSNR by more than 1 dB. Previous methods fail to reconstruct a high-fidelity 4D scene with sparse inputs. Sparse4DGS consistently achieves the best performance on three publicly available datasets. These results confirm the ability of Sparse4DGS to achieve photorealistic reconstruction even with sparse input frames.

\noindent\textbf{Qualitative results.} 
Fig.~\ref{fig:vis_blender_20} and Fig.~\ref{fig:vis_nerfds_20} present the visualization results of Sparse4DGS and other methods. The evaluation set consists of novel viewpoints (spatial interpolation) and novel temporal frames. Although CoRGS~\cite{zhang2025cor} exhibits a strong ability to synthesize novel views and novel frames from sparse inputs, it fails to effectively handle dynamic objects. Deformable3DGS~\cite{yang2024deformable} and 4DGaussians~\cite{wu20244d} experience significant performance drops, particularly in regions around object edges. With \textit{TACO} and \textit{TADR}, the Gaussians in Sparse4DGS are rigorously constrained to align with object edges, thereby preserving detailed structural information.

\subsection{Multi Frame Rate Application}
In this section, to evaluate our method in the real-world application, we test Sparse4DGS on the iPhone-4D dataset using input sequences at 5 FPS and 30 FPS. 
% \textbf{Quantitative results}
We present the quantitative results in Tab.~\ref{tab:comparison_all}. Our method consistently outperforms previous baselines for input sequences both at 5 FPS and 30 FPS. In real-world frame-rate applications, Sparse4DGS demonstrates superior reconstruction performance, particularly with extremely low-frame-rate input videos.  
% \textbf{Qualitative results}
The qualitative results are shown in Fig.~\ref{fig:vis_iphone_50}. Our method exhibits more accurate reconstruction of dynamic object edges. In the case of the ``\textit{MacBook}'', our method reconstructs a clear Apple logo, while other baselines suffer from blurry results. These results demonstrate that our method can achieve photo-realistic novel view and novel frame synthesis result with sparse input frame sequences. The results on iPhone-4D demonstrate the effectiveness of Sparse4DGS in real-world frame-rate applications. 
Detailed quantitative and qualitative results can be found in our supplementary materials.

\begin{figure}
    \centering
    \includegraphics[width=0.95\linewidth]{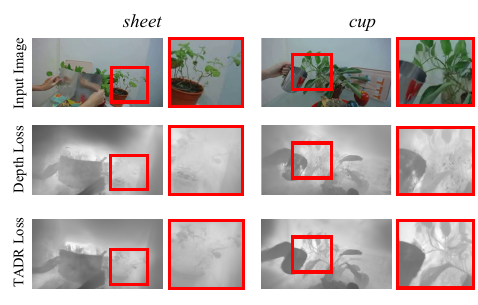}
   \caption{\textbf{Illustration of \textit{TADR}.} TADR demonstrates superior performance in capturing local depth variations.}
    \label{fig:tadr_l1}
\end{figure}

\begin{figure}
    \centering
    \includegraphics[width=0.92\linewidth]{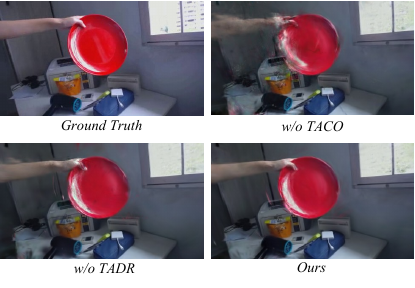}
   \caption{\textbf{Illustration of our architecture modules.} Removing \textit{TADR} or \textit{TACO} results in performance drop.}
    \label{fig:major_ablation_vis}
\end{figure}

\subsection{Ablation Study}
\label{sec:ablation_study}
We ablate the effectiveness of Sparse4DGS on NeRF-DS dataset with 20 input images. 

\noindent\textbf{Architecture modules.} In this section, we conduct a detailed analysis of the effectiveness of our proposed modules. The experimental results are summarized in Tab.~\ref{tab:major_ablation} and Fig.~\ref{fig:major_ablation_vis}. Removing \textit{TADR} and \textit{TACO} causes PSNR drops of 0.45 and 1.01, respectively, demonstrating their critical role in optimizing a dynamic scene.

\noindent\textbf{Selection of $\lambda_1$ and $\lambda_2$.} Hyperparameter tuning results in Tab.~\ref{tab:lambda_1} and Tab.~\ref{tab:lambda_2}. Setting $\lambda_1$ and $\lambda_2$ to 0.01 yields the best performance for Sparse4DGS.

\noindent\textbf{Noise terms of \textit{TACO}.} \textit{TACO} consists of two noise terms: $\epsilon_o$ and $\epsilon_{tex}$. Removing either causes a performance drop (Tab.~\ref{tab:taco_epsilon}). Without $\epsilon_{tex}$, Gaussians fail to concentrate in texture-rich areas, while $\epsilon_o$ helps converge ambiguous Gaussians that are too opaque for pruning.

\noindent\textbf{Ablation study on PCC.} Fig.~\ref{fig:edge_pcc_l1} shows the effectiveness of PCC. Replacing PCC loss in $L_{tex}$ and $L_{tadr}$ with L1 loss results in performance drops of 0.6 and 0.3. This is because L1 loss overlooks spatial inconsistencies in multi-view texture and depth maps (Tab.~\ref{tab:pcc_ablate}).

\noindent\textbf{Ablation study on texture-aware depth loss in TADR.} Unlike conventional depth regularization methods~\cite{he2025see, zhu2025fsgs, kumar2024few, gao2024relightable}, this work introduces a texture-aware depth loss~(Eq.\ref{eq:l_tadr}) in \textit{TADR}. We compare our method with a conventional depth loss, which directly calculates PCC between $D_{render}$ and $D_{dpt}$. As shown in Fig.~\ref{fig:tadr_l1} and Tab.~\ref{tab:tadr_detail}, both quantitative and qualitative results demonstrate the effectiveness of adopting texture in depth loss.

\section{Conclusion}
In this work, we introduce \textbf{Sparse4DGS}, the first method for dynamic scene reconstruction using sparse input frame sequences. We identify the challenges in optimizing deformation and canonical fields, particularly in regions with pronounced texture variations. To tackle this, we propose a texture-aware method (\textit{TADR} and \textit{TACO}) that aligns Gaussians with texture information. We evaluate \textbf{Sparse4DGS} on our iPhone-4D dataset with frame sequences captured at 5 FPS or 30 FPS, representing a significant advancement for real-world applications in dynamic scene reconstruction.

\section*{Acknowledgments}
This work was supported in part by the National Natural Science Foundation of China under Grants (No. 62206082, 422204, 62502135, 62402152, and 62406093), the Key
Research and Development Program of Zhejiang Province
(No. 2025C01026), the Zhejiang Provincial Natural Science Foundation of China under Grants
(No. LDT23F02025F02, LQ24F020032, LQN25F030014, and LQN25F020017). This research was also supported by the National College Student Innovation and Entrepreneurship Training Program of China under Grants (No. 202410336018 and 202510336016).

\bibliography{main}

\begin{thebibliography}{52}
\providecommand{\natexlab}[1]{#1}

\bibitem[{Amari(1993)}]{amari1993backpropagation}
Amari, S.-i. 1993.
\newblock Backpropagation and stochastic gradient descent method.
\newblock \emph{Neurocomputing}, 5(4-5): 185--196.

\bibitem[{Barron et~al.(2021)Barron, Mildenhall, Tancik, Hedman, Martin-Brualla, and Srinivasan}]{barron2021mip}
Barron, J.~T.; Mildenhall, B.; Tancik, M.; Hedman, P.; Martin-Brualla, R.; and Srinivasan, P.~P. 2021.
\newblock Mip-nerf: A multiscale representation for anti-aliasing neural radiance fields.
\newblock In \emph{Proceedings of the IEEE/CVF international conference on computer vision}, 5855--5864.

\bibitem[{Barron et~al.(2022)Barron, Mildenhall, Verbin, Srinivasan, and Hedman}]{barron2022mip}
Barron, J.~T.; Mildenhall, B.; Verbin, D.; Srinivasan, P.~P.; and Hedman, P. 2022.
\newblock Mip-nerf 360: Unbounded anti-aliased neural radiance fields.
\newblock In \emph{Proceedings of the IEEE/CVF conference on computer vision and pattern recognition}, 5470--5479.

\bibitem[{Brosse et~al.(2018)Brosse, Durmus, Moulines, and and}]{brosse2018promises}
Brosse, N.; Durmus, A.; Moulines, E.; and and. 2018.
\newblock The promises and pitfalls of stochastic gradient Langevin dynamics.
\newblock \emph{Advances in Neural Information Processing Systems}, 31.

\bibitem[{Charatan et~al.(2024)Charatan, Li, Tagliasacchi, and Sitzmann}]{charatan2024pixelsplat}
Charatan, D.; Li, S.~L.; Tagliasacchi, A.; and Sitzmann, V. 2024.
\newblock pixelsplat: 3d gaussian splats from image pairs for scalable generalizable 3d reconstruction.
\newblock In \emph{Proceedings of the IEEE/CVF Conference on Computer Vision and Pattern Recognition}, 19457--19467.

\bibitem[{Chen et~al.(2022)Chen, Xu, Geiger, Yu, and Su}]{chen2022tensorf}
Chen, A.; Xu, Z.; Geiger, A.; Yu, J.; and Su, H. 2022.
\newblock Tensorf: Tensorial radiance fields.
\newblock In \emph{European conference on computer vision}, 333--350. Springer.

\bibitem[{Chen et~al.(2025)Chen, Xu, Zheng, Zhuang, Pollefeys, Geiger, Cham, and Cai}]{chen2025mvsplat}
Chen, Y.; Xu, H.; Zheng, C.; Zhuang, B.; Pollefeys, M.; Geiger, A.; Cham, T.-J.; and Cai, J. 2025.
\newblock Mvsplat: Efficient 3d gaussian splatting from sparse multi-view images.
\newblock In \emph{European Conference on Computer Vision}, 370--386. Springer.

\bibitem[{Choi et~al.(2025)Choi, Song, Kim, Kim, and Do}]{choi2025click}
Choi, S.; Song, H.; Kim, J.; Kim, T.; and Do, H. 2025.
\newblock Click-gaussian: Interactive segmentation to any 3d gaussians.
\newblock In \emph{European Conference on Computer Vision}, 289--305. Springer.

\bibitem[{Cohen et~al.(2009)Cohen, Huang, Chen, Benesty, Benesty, Chen, Huang, and Cohen}]{cohen2009pearson}
Cohen, I.; Huang, Y.; Chen, J.; Benesty, J.; Benesty, J.; Chen, J.; Huang, Y.; and Cohen, I. 2009.
\newblock Pearson correlation coefficient.
\newblock \emph{Noise reduction in speech processing}, 1--4.

\bibitem[{Duan et~al.(2024)Duan, Wei, Dai, He, Chen, and Chen}]{duan20244d}
Duan, Y.; Wei, F.; Dai, Q.; He, Y.; Chen, W.; and Chen, B. 2024.
\newblock 4d gaussian splatting: Towards efficient novel view synthesis for dynamic scenes.
\newblock \emph{arXiv preprint arXiv:2402.03307}.

\bibitem[{Duisterhof et~al.(2023)Duisterhof, Mandi, Yao, Liu, Shou, Song, and Ichnowski}]{duisterhof2023md}
Duisterhof, B.~P.; Mandi, Z.; Yao, Y.; Liu, J.-W.; Shou, M.~Z.; Song, S.; and Ichnowski, J. 2023.
\newblock Md-splatting: Learning metric deformation from 4d gaussians in highly deformable scenes.
\newblock \emph{arXiv preprint arXiv:2312.00583}.

\bibitem[{Feng et~al.(2024)Feng, He, Wang, Yang, Li, Chen, Kuang, Fan, Jun et~al.}]{feng2024srgs}
Feng, X.; He, Y.; Wang, Y.; Yang, Y.; Li, W.; Chen, Y.; Kuang, Z.; Fan, J.; Jun, Y.; et~al. 2024.
\newblock Srgs: Super-resolution 3d gaussian splatting.
\newblock \emph{arXiv preprint arXiv:2404.10318}.

\bibitem[{Gao et~al.(2024)Gao, Gu, Lin, Li, Zhu, Cao, Zhang, and Yao}]{gao2024relightable}
Gao, J.; Gu, C.; Lin, Y.; Li, Z.; Zhu, H.; Cao, X.; Zhang, L.; and Yao, Y. 2024.
\newblock Relightable 3d gaussians: Realistic point cloud relighting with brdf decomposition and ray tracing.
\newblock In \emph{European Conference on Computer Vision}, 73--89. Springer.

\bibitem[{Gao et~al.(2025)Gao, Gu, Lin, Li, Zhu, Cao, Zhang, and Yao}]{gao2025relightable}
Gao, J.; Gu, C.; Lin, Y.; Li, Z.; Zhu, H.; Cao, X.; Zhang, L.; and Yao, Y. 2025.
\newblock Relightable 3d gaussians: Realistic point cloud relighting with brdf decomposition and ray tracing.
\newblock In \emph{European Conference on Computer Vision}, 73--89. Springer.

\bibitem[{He et~al.(2025)He, Xiao, Chan, Zuo, Xiao, and Lam}]{he2025see}
He, Z.; Xiao, Z.; Chan, K.-C.; Zuo, Y.; Xiao, J.; and Lam, K.-M. 2025.
\newblock See In Detail: Enhancing Sparse-view 3D Gaussian Splatting with Local Depth and Semantic Regularization.
\newblock \emph{arXiv preprint arXiv:2501.11508}.

\bibitem[{Hu et~al.(2025)Hu, Shi, Yang, Chen, Gu, Ding, He, and Fan}]{hu2025texture}
Hu, X.; Shi, C.; Yang, C.; Chen, M.; Gu, X.; Ding, J.; He, J.; and Fan, J. 2025.
\newblock Texture-aware 3D Gaussian Splatting for sparse view reconstructions.
\newblock \emph{Applied Soft Computing}, 113530.

\bibitem[{Huang et~al.(2024)Huang, Sun, Yang, Lyu, Cao, and Qi}]{huang2024sc}
Huang, Y.-H.; Sun, Y.-T.; Yang, Z.; Lyu, X.; Cao, Y.-P.; and Qi, X. 2024.
\newblock Sc-gs: Sparse-controlled gaussian splatting for editable dynamic scenes.
\newblock In \emph{Proceedings of the IEEE/CVF Conference on Computer Vision and Pattern Recognition}, 4220--4230.

\bibitem[{Jiang et~al.(2024)Jiang, Yu, Xie, Li, Feng, Wang, Li, Lau, Gao, Yang et~al.}]{jiang2024vr}
Jiang, Y.; Yu, C.; Xie, T.; Li, X.; Feng, Y.; Wang, H.; Li, M.; Lau, H.; Gao, F.; Yang, Y.; et~al. 2024.
\newblock Vr-gs: A physical dynamics-aware interactive gaussian splatting system in virtual reality.
\newblock In \emph{ACM SIGGRAPH 2024 Conference Papers}, 1--1.

\bibitem[{Kerbl et~al.(2023)Kerbl, Kopanas, Leimk{\"u}hler, and Drettakis}]{kerbl3Dgaussians}
Kerbl, B.; Kopanas, G.; Leimk{\"u}hler, T.; and Drettakis, G. 2023.
\newblock 3D Gaussian splatting for real-time radiance field rendering.
\newblock \emph{ACM Trans. Graph.}, 42(4): 139--1.

\bibitem[{Kheradmand et~al.(2024{\natexlab{a}})Kheradmand, Rebain, Sharma, Isack, Kar, Tagliasacchi, and Yi}]{kheradmand2024accelerating}
Kheradmand, S.; Rebain, D.; Sharma, G.; Isack, H.; Kar, A.; Tagliasacchi, A.; and Yi, K.~M. 2024{\natexlab{a}}.
\newblock Accelerating Neural Field Training via Soft Mining.
\newblock In \emph{Proceedings of the IEEE/CVF Conference on Computer Vision and Pattern Recognition}, 20071--20080.

\bibitem[{Kheradmand et~al.(2024{\natexlab{b}})Kheradmand, Rebain, Sharma, Sun, Tseng, Isack, Kar, Tagliasacchi, and Yi}]{kheradmand20243d}
Kheradmand, S.; Rebain, D.; Sharma, G.; Sun, W.; Tseng, J.; Isack, H.; Kar, A.; Tagliasacchi, A.; and Yi, K.~M. 2024{\natexlab{b}}.
\newblock 3D Gaussian Splatting as Markov Chain Monte Carlo.
\newblock \emph{arXiv preprint arXiv:2404.09591}.

\bibitem[{Kumar and Vats(2024)}]{kumar2024few}
Kumar, R.; and Vats, V. 2024.
\newblock Few-shot Novel View Synthesis using Depth Aware 3D Gaussian Splatting.
\newblock \emph{arXiv preprint arXiv:2410.11080}.

\bibitem[{Lee et~al.(2025)Lee, Lee, Sun, Ali, and Park}]{lee2025deblurring}
Lee, B.; Lee, H.; Sun, X.; Ali, U.; and Park, E. 2025.
\newblock Deblurring 3d gaussian splatting.
\newblock In \emph{European Conference on Computer Vision}, 127--143. Springer.

\bibitem[{Li et~al.(2024)Li, Zhang, Bai, Zheng, Ning, Zhou, and Gu}]{li2024dngaussian}
Li, J.; Zhang, J.; Bai, X.; Zheng, J.; Ning, X.; Zhou, J.; and Gu, L. 2024.
\newblock Dngaussian: Optimizing sparse-view 3d gaussian radiance fields with global-local depth normalization.
\newblock In \emph{Proceedings of the IEEE/CVF Conference on Computer Vision and Pattern Recognition}, 20775--20785.

\bibitem[{Liang et~al.(2023)Liang, Khan, Li, Nguyen-Phuoc, Lanman, Tompkin, and Xiao}]{liang2023gaufre}
Liang, Y.; Khan, N.; Li, Z.; Nguyen-Phuoc, T.; Lanman, D.; Tompkin, J.; and Xiao, L. 2023.
\newblock Gaufre: Gaussian deformation fields for real-time dynamic novel view synthesis.
\newblock \emph{arXiv preprint arXiv:2312.11458}.

\bibitem[{Mildenhall et~al.(2021)Mildenhall, Srinivasan, Tancik, Barron, Ramamoorthi, and Ng}]{mildenhall2021nerf}
Mildenhall, B.; Srinivasan, P.~P.; Tancik, M.; Barron, J.~T.; Ramamoorthi, R.; and Ng, R. 2021.
\newblock Nerf: Representing scenes as neural radiance fields for view synthesis.
\newblock \emph{Communications of the ACM}, 65(1): 99--106.

\bibitem[{M{\"u}ller et~al.(2022)M{\"u}ller, Evans, Schied, and Keller}]{muller2022instant}
M{\"u}ller, T.; Evans, A.; Schied, C.; and Keller, A. 2022.
\newblock Instant neural graphics primitives with a multiresolution hash encoding.
\newblock \emph{ACM transactions on graphics (TOG)}, 41(4): 1--15.

\bibitem[{Paliwal et~al.(2025)Paliwal, Ye, Xiong, Kotovenko, Ranjan, Chandra, and Kalantari}]{paliwal2025coherentgs}
Paliwal, A.; Ye, W.; Xiong, J.; Kotovenko, D.; Ranjan, R.; Chandra, V.; and Kalantari, N.~K. 2025.
\newblock Coherentgs: Sparse novel view synthesis with coherent 3d gaussians.
\newblock In \emph{European Conference on Computer Vision}, 19--37. Springer.

\bibitem[{Park et~al.(2021)Park, Sinha, Hedman, Barron, Bouaziz, Goldman, Martin-Brualla, and Seitz}]{park2021hypernerf}
Park, K.; Sinha, U.; Hedman, P.; Barron, J.~T.; Bouaziz, S.; Goldman, D.~B.; Martin-Brualla, R.; and Seitz, S.~M. 2021.
\newblock Hypernerf: A higher-dimensional representation for topologically varying neural radiance fields.
\newblock \emph{arXiv preprint arXiv:2106.13228}.

\bibitem[{Pumarola et~al.(2021)Pumarola, Corona, Pons-Moll, and Moreno-Noguer}]{pumarola2021d}
Pumarola, A.; Corona, E.; Pons-Moll, G.; and Moreno-Noguer, F. 2021.
\newblock D-nerf: Neural radiance fields for dynamic scenes.
\newblock In \emph{Proceedings of the IEEE/CVF Conference on Computer Vision and Pattern Recognition}, 10318--10327.

\bibitem[{Qin et~al.(2024)Qin, Li, Zhou, Wang, and Pfister}]{qin2024langsplat}
Qin, M.; Li, W.; Zhou, J.; Wang, H.; and Pfister, H. 2024.
\newblock Langsplat: 3d language gaussian splatting.
\newblock In \emph{Proceedings of the IEEE/CVF Conference on Computer Vision and Pattern Recognition}, 20051--20060.

\bibitem[{Ranftl et~al.(2021)Ranftl, Bochkovskiy, Koltun, and and}]{ranftl2021vision}
Ranftl, R.; Bochkovskiy, A.; Koltun, V.; and and. 2021.
\newblock Vision transformers for dense prediction.
\newblock In \emph{Proceedings of the IEEE/CVF international conference on computer vision}, 12179--12188.

\bibitem[{Rombach et~al.(2022)Rombach, Blattmann, Lorenz, Esser, and Ommer}]{rombach2021highresolution}
Rombach, R.; Blattmann, A.; Lorenz, D.; Esser, P.; and Ommer, B. 2022.
\newblock High-resolution image synthesis with latent diffusion models.
\newblock In \emph{Proceedings of the IEEE/CVF conference on computer vision and pattern recognition}, 10684--10695.

\bibitem[{Saroha et~al.(2025)Saroha, Hofherr, Gladkova, Curreli, Litany, and Cremers}]{saroha2025zdyss}
Saroha, A.; Hofherr, F.; Gladkova, M.; Curreli, C.; Litany, O.; and Cremers, D. 2025.
\newblock ZDySS--Zero-Shot Dynamic Scene Stylization using Gaussian Splatting.
\newblock \emph{arXiv preprint arXiv:2501.03875}.

\bibitem[{Shan et~al.(2025)Shan, Cai, Hsieh, Cheng, and Wang}]{shan2025deformable}
Shan, J.; Cai, Z.; Hsieh, C.-T.; Cheng, S.~S.; and Wang, H. 2025.
\newblock Deformable Gaussian Splatting for Efficient and High-Fidelity Reconstruction of Surgical Scenes.
\newblock \emph{arXiv preprint arXiv:2501.01101}.

\bibitem[{Shi et~al.(2025{\natexlab{a}})Shi, Chen, Mao, Yang, Hu, Ding, and Yu}]{shi2025realm}
Shi, C.; Chen, M.; Mao, Y.; Yang, C.; Hu, X.; Ding, J.; and Yu, Z. 2025{\natexlab{a}}.
\newblock REALM: An MLLM-Agent Framework for Open World 3D Reasoning Segmentation and Editing on Gaussian Splatting.
\newblock \emph{arXiv preprint arXiv:2510.16410}.

\bibitem[{Shi et~al.(2025{\natexlab{b}})Shi, Yang, Hu, Yang, Ding, and Tan}]{shi2025mmgs}
Shi, C.; Yang, C.; Hu, X.; Yang, Y.; Ding, J.; and Tan, M. 2025{\natexlab{b}}.
\newblock MMGS: Multi-Model Synergistic Gaussian Splatting for Sparse View Synthesis.
\newblock \emph{Image and Vision Computing}, 158: 105512.

\bibitem[{Tomasi and Kanade(1992)}]{tomasi1992shape}
Tomasi, C.; and Kanade, T. 1992.
\newblock Shape and motion from image streams under orthography: a factorization method.
\newblock \emph{International journal of computer vision}, 9: 137--154.

\bibitem[{Ullman(1979)}]{ullman1979interpretation}
Ullman, S. 1979.
\newblock The interpretation of structure from motion.
\newblock \emph{Proceedings of the Royal Society of London. Series B. Biological Sciences}, 203(1153): 405--426.

\bibitem[{Wang et~al.(2004)Wang, Bovik, Sheikh, and Simoncelli}]{wang2004image}
Wang, Z.; Bovik, A.~C.; Sheikh, H.~R.; and Simoncelli, E.~P. 2004.
\newblock Image quality assessment: from error visibility to structural similarity.
\newblock \emph{IEEE transactions on image processing}, 13(4): 600--612.

\bibitem[{Welling and Teh(2011)}]{welling2011bayesian}
Welling, M.; and Teh, Y.~W. 2011.
\newblock Bayesian learning via stochastic gradient Langevin dynamics.
\newblock In \emph{Proceedings of the 28th international conference on machine learning (ICML-11)}, 681--688. Citeseer.

\bibitem[{Wu et~al.(2024{\natexlab{a}})Wu, Yi, Fang, Xie, Zhang, Wei, Liu, Tian, and Wang}]{wu20244d}
Wu, G.; Yi, T.; Fang, J.; Xie, L.; Zhang, X.; Wei, W.; Liu, W.; Tian, Q.; and Wang, X. 2024{\natexlab{a}}.
\newblock 4d gaussian splatting for real-time dynamic scene rendering.
\newblock In \emph{Proceedings of the IEEE/CVF Conference on Computer Vision and Pattern Recognition}, 20310--20320.

\bibitem[{Wu et~al.(2024{\natexlab{b}})Wu, Zhang, Chen, Fan, Yan, and Zuo}]{wu2024deblur4dgs}
Wu, R.; Zhang, Z.; Chen, M.; Fan, X.; Yan, Z.; and Zuo, W. 2024{\natexlab{b}}.
\newblock Deblur4DGS: 4D Gaussian Splatting from Blurry Monocular Video.
\newblock \emph{arXiv preprint arXiv:2412.06424}.

\bibitem[{Yan, Li, and Lee(2023)}]{yan2023nerf}
Yan, Z.; Li, C.; and Lee, G.~H. 2023.
\newblock Nerf-ds: Neural radiance fields for dynamic specular objects.
\newblock In \emph{Proceedings of the IEEE/CVF Conference on Computer Vision and Pattern Recognition}, 8285--8295.

\bibitem[{Yang et~al.(2024{\natexlab{a}})Yang, Li, Fang, Liang, Xie, Zhang, Shen, and Tian}]{yang2024gaussianobject}
Yang, C.; Li, S.; Fang, J.; Liang, R.; Xie, L.; Zhang, X.; Shen, W.; and Tian, Q. 2024{\natexlab{a}}.
\newblock Gaussianobject: High-quality 3d object reconstruction from four views with gaussian splatting.
\newblock \emph{ACM Transactions on Graphics (TOG)}, 43(6): 1--13.

\bibitem[{Yang et~al.(2024{\natexlab{b}})Yang, Gao, Zhou, Jiao, Zhang, and Jin}]{yang2024deformable}
Yang, Z.; Gao, X.; Zhou, W.; Jiao, S.; Zhang, Y.; and Jin, X. 2024{\natexlab{b}}.
\newblock Deformable 3d gaussians for high-fidelity monocular dynamic scene reconstruction.
\newblock In \emph{Proceedings of the IEEE/CVF Conference on Computer Vision and Pattern Recognition}, 20331--20341.

\bibitem[{Yang et~al.(2024{\natexlab{c}})Yang, Pan, Zhu, Zhang, Jiang, and Torr}]{yang20244d}
Yang, Z.; Pan, Z.; Zhu, X.; Zhang, L.; Jiang, Y.-G.; and Torr, P.~H. 2024{\natexlab{c}}.
\newblock 4D Gaussian Splatting: Modeling Dynamic Scenes with Native 4D Primitives.
\newblock \emph{arXiv preprint arXiv:2412.20720}.

\bibitem[{Yang et~al.(2023)Yang, Yang, Pan, and Zhang}]{yang2023real}
Yang, Z.; Yang, H.; Pan, Z.; and Zhang, L. 2023.
\newblock Real-time photorealistic dynamic scene representation and rendering with 4d gaussian splatting.
\newblock \emph{arXiv preprint arXiv:2310.10642}.

\bibitem[{Yu et~al.(2021)Yu, Fridovich-Keil, Tancik, Chen, Recht, and Kanazawa}]{yu2021plenoxels}
Yu, A.; Fridovich-Keil, S.; Tancik, M.; Chen, Q.; Recht, B.; and Kanazawa, A. 2021.
\newblock Plenoxels: Radiance fields without neural networks.
\newblock \emph{arXiv preprint arXiv:2112.05131}, 2(3): 6.

\bibitem[{Zhang et~al.(2025)Zhang, Li, Yu, Huang, Gu, Zheng, and Bai}]{zhang2025cor}
Zhang, J.; Li, J.; Yu, X.; Huang, L.; Gu, L.; Zheng, J.; and Bai, X. 2025.
\newblock CoR-GS: sparse-view 3D Gaussian splatting via co-regularization.
\newblock In \emph{European Conference on Computer Vision}, 335--352. Springer.

\bibitem[{Zhang et~al.(2018)Zhang, Isola, Efros, Shechtman, and Wang}]{zhang2018unreasonable}
Zhang, R.; Isola, P.; Efros, A.~A.; Shechtman, E.; and Wang, O. 2018.
\newblock The unreasonable effectiveness of deep features as a perceptual metric.
\newblock In \emph{Proceedings of the IEEE conference on computer vision and pattern recognition}, 586--595.

\bibitem[{Zhu et~al.(2025)Zhu, Fan, Jiang, and Wang}]{zhu2025fsgs}
Zhu, Z.; Fan, Z.; Jiang, Y.; and Wang, Z. 2025.
\newblock Fsgs: Real-time few-shot view synthesis using gaussian splatting.
\newblock In \emph{European conference on computer vision}, 145--163. Springer.

\end{thebibliography}

\end{document}